%% file: main.tex
\documentclass[conference]{IEEEtran}
\makeatletter
\def\ps@IEEEtitlepagestyle{%
  \def\@oddfoot{\mycopyrightnotice}%
  \def\@oddhead{\hbox{}\@IEEEheaderstyle\leftmark\hfil\thepage}\relax
  \def\@evenhead{\@IEEEheaderstyle\thepage\hfil\leftmark\hbox{}}\relax
  \def\@evenfoot{}%
}
\def\mycopyrightnotice{%
  \begin{minipage}{\textwidth}
  \scriptsize
  \copyright~2023 IEEE. Personal use of this material is permitted. Permission from IEEE must be obtained for all other uses, in any current or future media, including reprinting/republishing this material for advertising or promotional purposes, creating new collective works, for resale or redistribution to servers or lists, or reuse of any copyrighted component of this work in other works. 
  This work has been accepted at The 33rd International Conference on Field-Programmable Logic and Applications (FPL), 2023.
  \end{minipage}
}
\makeatother

\usepackage{booktabs} 
\usepackage{enumerate}
\usepackage{graphicx}
\usepackage{amsmath}
\usepackage{multirow}
\usepackage{footnote}
\usepackage{threeparttable}
\usepackage{amsmath}
\usepackage{textcomp}
\usepackage{balance}
\usepackage{hyperref}
\usepackage[table,xcdraw]{xcolor}

\newcommand{\secref}[1]{Section~\ref{#1}}
\newcommand{\tabref}[1]{Table~\ref{#1}}

\usepackage[square, comma, numbers]{natbib}

\usepackage{listings}

\usepackage[colorinlistoftodos]{todonotes} 

\usepackage[caption=false]{subfig}

\definecolor{codegreen}{rgb}{0,0.6,0}
\definecolor{codegray}{rgb}{0.5,0.5,0.5}
\definecolor{codepurple}{rgb}{0.58,0,0.82}
\definecolor{backcolour}{rgb}{0.95,0.95,0.92}

\lstdefinestyle{mystyle}{
  backgroundcolor=\color{backcolour}, 
  commentstyle=\color{codegreen},
  keywordstyle=\color{magenta},
  numberstyle=\tiny\color{codegray},
  stringstyle=\color{codepurple},
  basicstyle=\ttfamily\scriptsize
  breakatwhitespace=false,    
  language=Python,
  breaklines=true,                 
  captionpos=b,                    
  keepspaces=true,                 
  numbers=left,                    
  numbersep=5pt,                  
  showspaces=false,                
  showstringspaces=false,
  showtabs=false,                  
  tabsize=2,
  float=tp
}
\begin{document}
%

\title{MetaML: Automating Customizable Cross-Stage Design-Flow for Deep Learning Acceleration}

\author{
\IEEEauthorblockN{
Zhiqiang Que, 
Shuo Liu, 
Markus Rognlien, 
Ce Guo, 
Jose~G.~F.~Coutinho, 
Wayne Luk
}

\IEEEauthorblockA{Department of Computing, Imperial College London, UK. \{z.que, c.guo, jgfc, w.luk\}@imperial.ac.uk }

}

\maketitle

\input{sections/abstract.tex}

\input{sections/intro.tex}

\input{sections/related_work.tex}
\input{sections/approach.tex}

\input{sections/evaluation.tex}
\input{sections/conclusion.tex}

\footnotesize
\bibliographystyle{IEEEtran}
\bibliography{main-bibliography}


\end{document}

%% file: sections/abstract.tex
\begin{abstract}


This paper introduces a novel optimization framework for deep neural network (DNN) hardware accelerators, enabling the rapid development of customized and automated design flows. More specifically, our approach aims to automate the selection and configuration of low-level optimization techniques, encompassing DNN and FPGA low-level optimizations. We introduce novel optimization and transformation tasks for building design-flow architectures, which are highly customizable and flexible, thereby enhancing the performance and efficiency of DNN accelerators. Our results demonstrate considerable reductions of up to 92\% in DSP usage and 89\% in LUT usage for two networks, while maintaining accuracy and eliminating the need for human effort or domain expertise. In comparison to state-of-the-art approaches, our design achieves higher accuracy and utilizes three times fewer DSP resources, underscoring the advantages of our proposed framework.

\end{abstract}

%% file: sections/intro.tex

\section{Introduction}


The field of deep learning has witnessed unprecedented growth in recent years, driven by the increasing demand for efficient and high-performance applications~\cite{lecun2015deep}. Consequently, FPGA-based deep neural network (DNN) accelerator design and optimization have gained significant attention~\cite{zhang2015optimizing, que2021recurrent, zhang2020dnnexplorer}. The development of an efficient FPGA-based DNN design requires a diverse skill set that combines expertise in machine learning with low-level knowledge of the target hardware architecture~\cite{sze2017efficient}. Optimizing these DNN designs is a complex process, as it involves balancing competing objectives. On one hand, high accuracy during inference is crucial from an application perspective. On the other hand, the design must be optimized for the underlying hardware architecture, meeting power, latency and throughput requirements while fitting into the FPGA device~\cite{coelho2021automatic, que2021accelerating}. Achieving an optimal balance between these conflicting objectives requires careful consideration and effective optimization strategies~\cite{wang2019deep}. 

In this paper, we focus on the key challenge of codifying optimization strategies that automate the selection and configuration of low-level optimization techniques covering multiple abstraction levels from DNN optimizations to FPGA optimizations. Achieving optimal results for a given problem is complex~\cite{deng2020model}. It is challenging to identify the most effective combination, order and tuning of optimization techniques to ensure optimal outcomes~\cite{zhang2020dnnexplorer, xu2020autodnnchip}.  In order to address this challenge, we propose the following contributions:
\begin{enumerate}
\item A co-optimization framework for FPGA-based DNN accelerators, which includes novel building tasks that enable rapid development of customized design flows, automating the entire design iteration process (Section~\ref{sec:approach}).

\item A library of reusable optimization and transformation tasks designed to be customizable and flexible, and that can be easily integrated into our co-optimization framework. Some of the tasks in our library are specific to certain applications and/or target technologies, while others are agnostic, providing versatility and adaptability to the framework (Section~\ref{sec:pipeblock}).

\item The evaluation of the proposed framework using multiple benchmarks and different optimization strategies. This evaluation provides insights into the effectiveness of the framework and its optimization modules under different scenarios (Section~\ref{sec:evaluation}).
\end{enumerate}

We present related work in~\secref{sec:related_work} and conclude this paper and present future work in~\secref{sec:conclusion}.

 

%% file: sections/related_work.tex
\section{Related Work}\label{sec:related_work}

The field of FPGA-based DNN acceleration has rapidly expanded, leading to the development of numerous optimization techniques and tools. 
Several co-optimization techniques have been proposed that optimize both algorithm and hardware stages for DNNs on FPGAs, as discussed in various papers~\cite{yang2019synetgy, hao2019fpga,  hao2020effective, jiang2020hardware, dong2021hao, fan2022algorithm, zhang2022algorithm} and in hardware-aware neural architecture search studies like~\cite{ney2021half, abdelfattah2020best}. These optimization strategies are often coupled with design space exploration, but are typically hardcoded and cannot be easily changed or customized.
Other approaches offer end-to-end software frameworks, such as Xilinx's Vitis AI~\cite{kathail2020xilinx} and Intel's OpenVINO~\cite{OpenVINO2023},  which optimize DNNs with pre-built optimizations for deployment on specific target technologies. 
Moreover, frameworks, such as FINN~\cite{umuroglu2017finn}, HLS4ML~\cite{duarte2018fast}, and fpgaConvNet~\cite{venieris2016fpgaconvnet}, provide optimized hardware building blocks for FPGA-based DNN accelerators. 
However, they do not support automated cross-stage optimization strategies. 

Our framework addresses the limitations of current optimization techniques for deep neural networks (DNNs) and hardware by supporting the design of fully automated optimization flows. These flows can be described by reusable tasks that are specific to the target platform as well as ones that can be used across different platforms.
Additionally, our approach seamlessly integrates both new and existing optimization techniques, catering to different levels of abstraction. For instance, it includes graph optimizations for neural networks and source-to-source optimizations for HLS C++ as described in~\cite{fccm20_artisan}.


%% file: sections/approach.tex
\section{Our Approach} \label{sec:approach}

This paper introduces a novel framework that simplifies the development of customizable design flows for optimizing deep neural networks (DNNs) on FPGA platforms. A design flow consists of a series of tasks that aim to convert a high-level specification into a final hardware design. Typically, a design flow is implemented as a multi-stage pipeline, where each stage operates on a specific model abstraction. As the pipeline progresses, the model abstraction is gradually refined and optimized, taking into account the key features of the target device. For instance, to illustrate the process, let us consider a DNN model described in Tensorflow. The framework utilizes the HLS4ML tool to translate this model into a C++ HLS model in the initial stage. Subsequently, the resulting C++ HLS model can be further transformed into a Register Transfer Level (RTL) model using Vivado HLS.

Our approach enables the optimization strategies to be codified, automating the selection and configuration of DNN and hardware optimizations. It is designed to fulfill the following requirements:

\textbf{Customizable}: Users have the freedom to customize, extend, or modify the design-flow to meet specific needs and support experimentation. They can select a set of design-flow tasks, that implement specific optimizations or transformations, arrange them in a desired order, and fine-tune their parameters to create a specific optimization flow. Additionally, users can develop their own tasks and integrate them into the design-flow.

\textbf{Cross-stage}: This refers to the breadth of optimizations and the ability to target multiple levels of abstraction, typically associated with different stages of a design-flow. More specifically, optimizations performed at the neural network level operate at the graph-level, while optimizations at the HLS C++ level involve source-level transformations.

\begin{figure}[tp]
   \centering
    \includegraphics[width=0.80\linewidth]{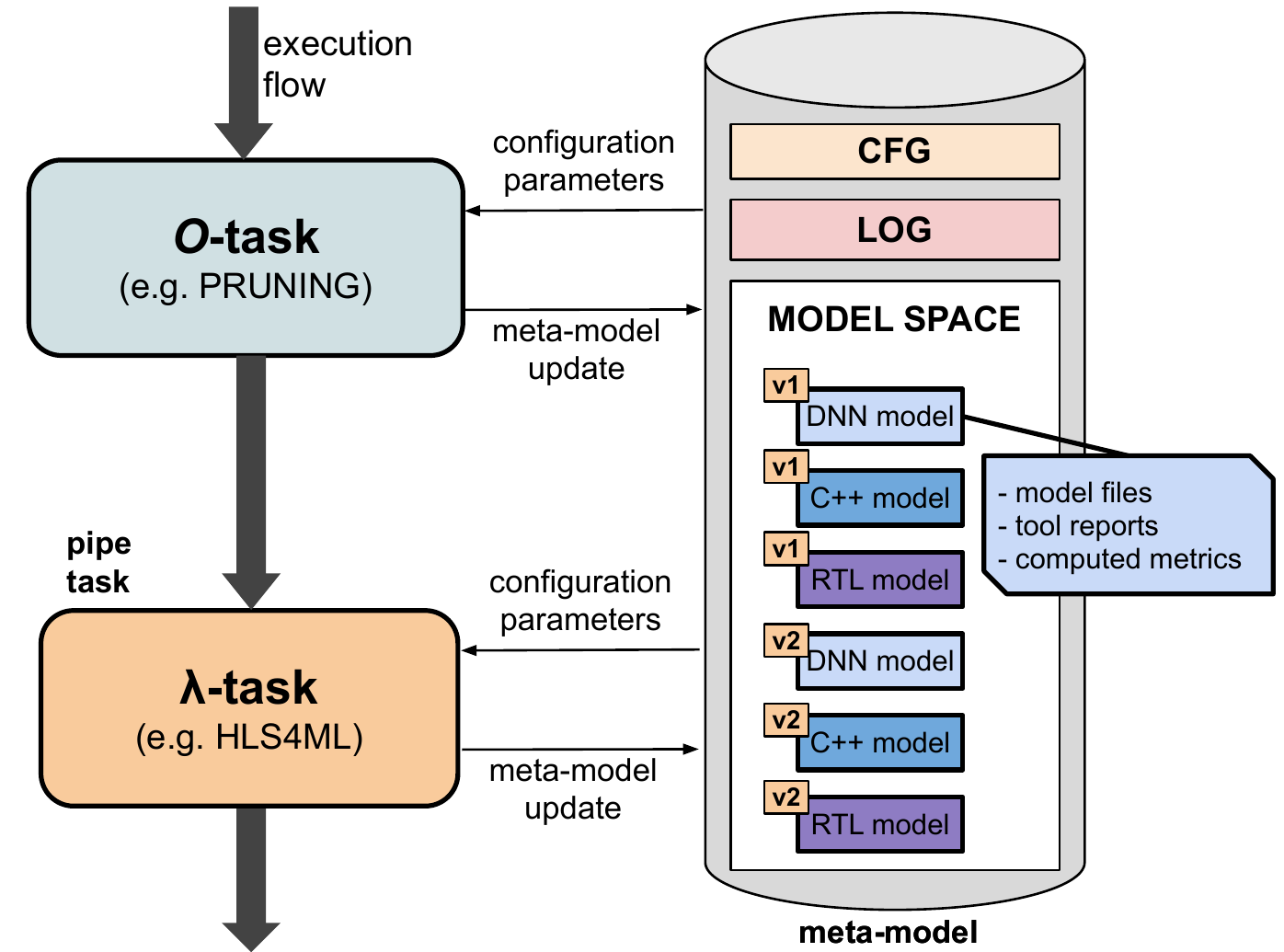}
  \caption{A connection between an $O$-task and a $\lambda$-task.  Each connection defines a unidirectional flow between a source and a target task.}
  \label{fig:pipeblocks} 
\end{figure}

Using our framework, design flows are programmatically generated and consist of two types of components (Fig.~\ref{fig:pipeblocks}): the pipe task and the meta-model. The pipe task serves as the basic unit of the design flow, executing specific optimizations or transformations. By interconnecting these tasks, we construct a complete design flow. The architecture of the design-flow is depicted as a cyclic directed graph where nodes symbolize tasks and edges signify dependencies between tasks, denoting the need to complete one task before initiating another.

The meta-model, on the other hand, serves as a shared space for storing the states of the design flow. This model consists of three sections: configuration, log, and model space. The configuration section (CFG) acts as a key-value store, holding the parameters of all pipe tasks in the design flow.  The log section (LOG) records the runtime execution trace, aiding in debugging. Furthermore, the model space can store the generated models obtained during the execution of the design flow. In the example presented in Fig.~\ref{fig:pipeblocks}, six models are stored, covering DNN, HLS C++, and RTL abstraction levels. Each model includes supporting files, tool reports, and computed metrics. 

\input{tables/pipeblocks03.tex}

\section{Reusable Pipe Tasks} \label{sec:pipeblock}


Table~\ref{table:pipeblocks} provides a list of pipe tasks that are currently implemented, along with their roles, multiplicity, and parameters. The multiplicity indicates the number of input and output connections that a task can handle. We consider two types of tasks:

\begin{itemize}

\item \textbf{$O$-task}: These are self-contained optimization tasks that enhance a given model based on specific objectives and constraints. Our current pipe task repository includes PRUNING and SCALING, which are implemented using the Keras API (version 2.9.0), and QUANTIZATION, which employs C++ source-to-source transformations via the Artisan framework~\cite{fccm20_artisan};

\item \textbf{$\lambda$-task}: These tasks perform functional transformations on the model space, such as compilation and synthesis. Examples include HLS4ML (version 0.6.0), which translates a DNN model into an HLS C++ model, and Vivado HLS (version 20.1), which translates an HLS C++ model into an RTL model.

\end{itemize}
Our framework is customizable. Different parameters (see Table~\ref{table:pipeblocks}) can be used to customise a design flow, and new design-flows can be built from existing ones to tailor to specific needs.




%% file: tables/pipeblocks03.tex
 \begin{table}[pb]
\centering
\caption{ A list of implemented pipe tasks.}
\label{table:pipeblocks}
\scalebox{.90}{
\begin{tabular}{l|c|c|l}
\hline
\rowcolor[HTML]{C0C0C0} 
\textbf{Type} &
  \textbf{Role} &
  \textbf{Multiplicity} &
  \textbf{Parameters}\\ \hline



HLS4ML  
  & $\lambda$ 
  & 1-to-1 & 
       \begin{tabular}[c]{@{}l@{}}
        default\_precision \\ 
        IOType \\
        FPGA\_part\_number \\
        clock\_period \\
        test\_dataset 
        \end{tabular}\\ \hline
        
VIVADO-HLS 
  & $\lambda$ 
  & 1-to-1 & 
       project\_dir \\ \hline

\begin{tabular}[c]{@{}l@{}}
KERAS-MODEL\\-GEN 
\end{tabular}
  & $\lambda$ 
  & 0-to-1 & 
       \begin{tabular}[c]{@{}l@{}}
        train\_en \\
        train\_test\_dataset\\
        train\_epochs
        \end{tabular}
  \\ \hline

PRUNING &
  $O$ &
  1-to-1 &
\begin{tabular}[c]{@{}l@{}}
        tolerate\_acc\_loss ($\alpha_p$) \\ 
        pruning\_rate\_thresh ($\beta_p$)\\
        train\_test\_dataset\\
        train\_epochs
        \end{tabular} 
        \\ \hline
SCALING         
  & $O$ 
  & 1-to-1 & 
       \begin{tabular}[c]{@{}l@{}}
        default\_scale\_factor \\ 
        tolerate\_acc\_loss ($\alpha_s$)\\ 
        scale\_auto \\
        max\_trials\_num \\ 
        train\_test\_dataset\\
        train\_epochs
        \end{tabular} \\ \hline

QUANTIZATION         
  & $O$ 
  & 1-to-1 & 
      \begin{tabular}[c]{@{}l@{}}
        tolerate\_acc\_loss ($\alpha_q$)\\ 
        train\_test\_dataset
        \end{tabular} \\ \hline

\end{tabular}
}
\end{table}

%% file: sections/evaluation.tex

\section{Evaluation}\label{sec:evaluation}
In this section, we demonstrate how optimization strategies can be built by revising design-flow architectures, combining and reusing pipe tasks, and modifying their configuration. 



\subsection{Experimental Setup}

Experiments were conducted in Python 3.9.15 with benchmark workloads from typical DNN applications, including jet identification~\cite{duarte2018fast, moreno2020jedi} (Jet-DNN), image classification using VGG7~\cite{simonyan2014very} and ResNet9~\cite{he2016deep} networks. The datasets used are: Jet-HLF~\cite{duarte2018fast}, MNIST~\cite{lecun1998gradient} and SVHN~\cite{netzer2011reading}, respectively. The jet identification task targeted FPGA-based CERN Large Hadron Collider (LHC) triggers with a 40 MHz input rate and a response latency of less than 1 microsecond. Default frequencies were 100MHz for Zynq 7020 and 200MHz for Alveo U250 and VU9P, and the HLS4ML task used 18-bit fixed-point precision with 8 integer bits.



\subsection{Optimization Strategies}\label{sec:single_opt}

In this subsection, we initially focus on three strategies, each supported by a single $O$-task. Then, we shift our focus towards combined strategies that utilize multiple $O$-tasks.

\begin{figure} 
   \centering
   \hspace*{\fill}
  \subfloat[]{%
    \includegraphics[width=0.44\linewidth]{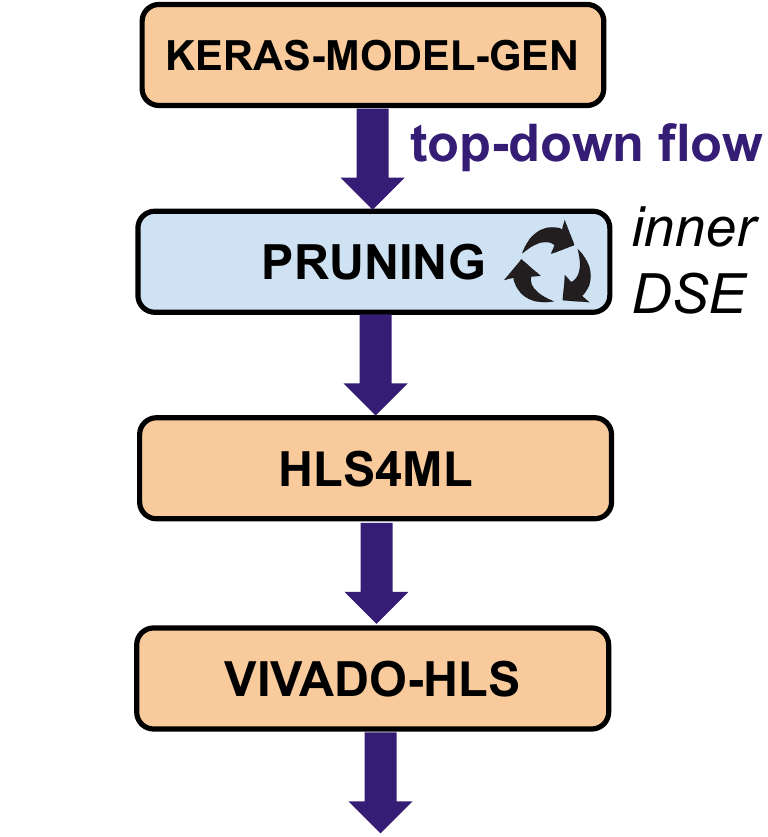}}
   \hspace*{\fill}
   \subfloat[]{%
    \includegraphics[width=0.28\linewidth]{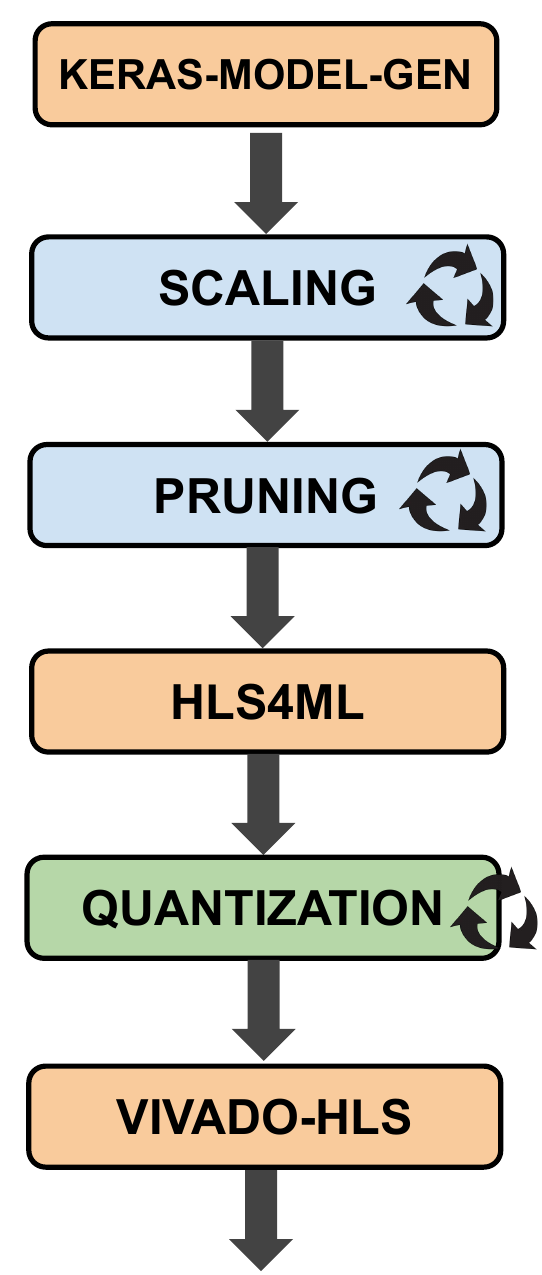}}
  \hspace*{\fill}
   \subfloat[]{%
    \includegraphics[width=0.28\linewidth]{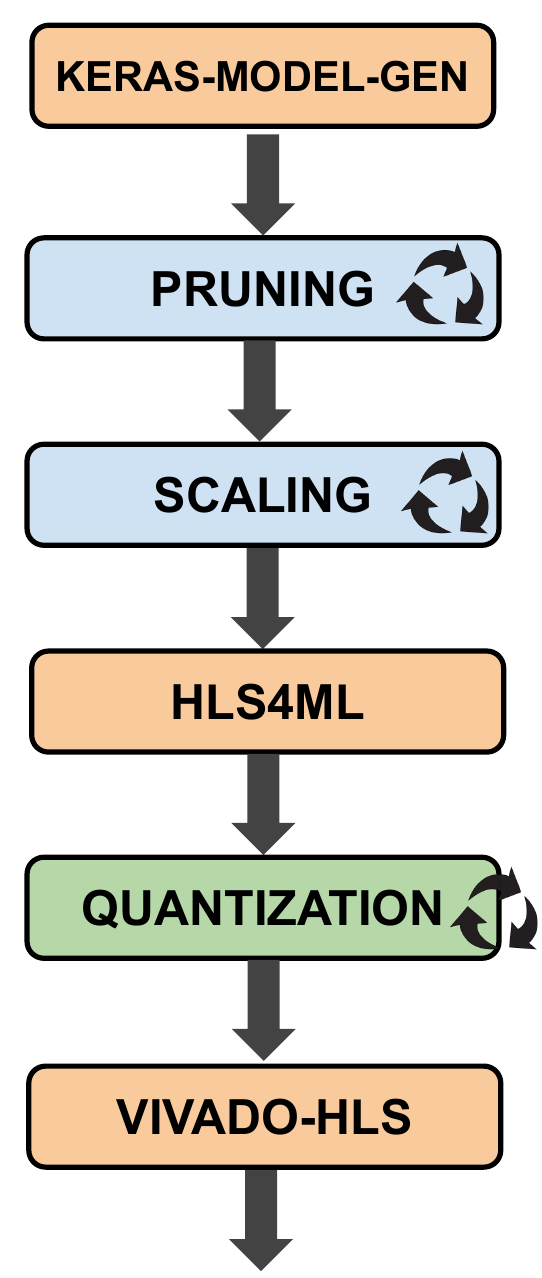}} 
 \hspace*{\fill}
  \caption{(a) Pruning strategy. 
  (b) The combined strategy of scaling, pruning and quantization. 
  (c) The combined strategy with a different order of $O$-tasks.  
  }
  \label{fig:design_flow} 
\vspace{0.2cm}
\end{figure}

\textbf{Pruning strategy}. 
Pruning is a technique that improves the performance and efficiency of neural networks by removing insignificant weights.   
Our framework includes an implementation of this technique through the PRUNING $O$-task. A design-flow which employs the PRUNING $O$-task is illustrated in Fig.~\ref{fig:design_flow}(a). This optimization task gradually zeroes out weights during training to create a more compact and efficient network while maintaining accuracy. In addition, it supports auto-pruning, which automatically determines the highest pruning rate while maintaining a given level of accuracy loss. Formally, the objective of this $O$-task is defined as:
\begin{equation}
\begin{footnotesize}
\begin{aligned}
& \underset{}{\text{maximum}}
& & Pruning\_rate \\
& \text{subject to}
& & Accuracy\_loss(Pruning\_rate) \leq \alpha_p 
\end{aligned}
\end{footnotesize}
\end{equation}

\begin{figure}
\begin{center}
\includegraphics[width=0.95\linewidth]{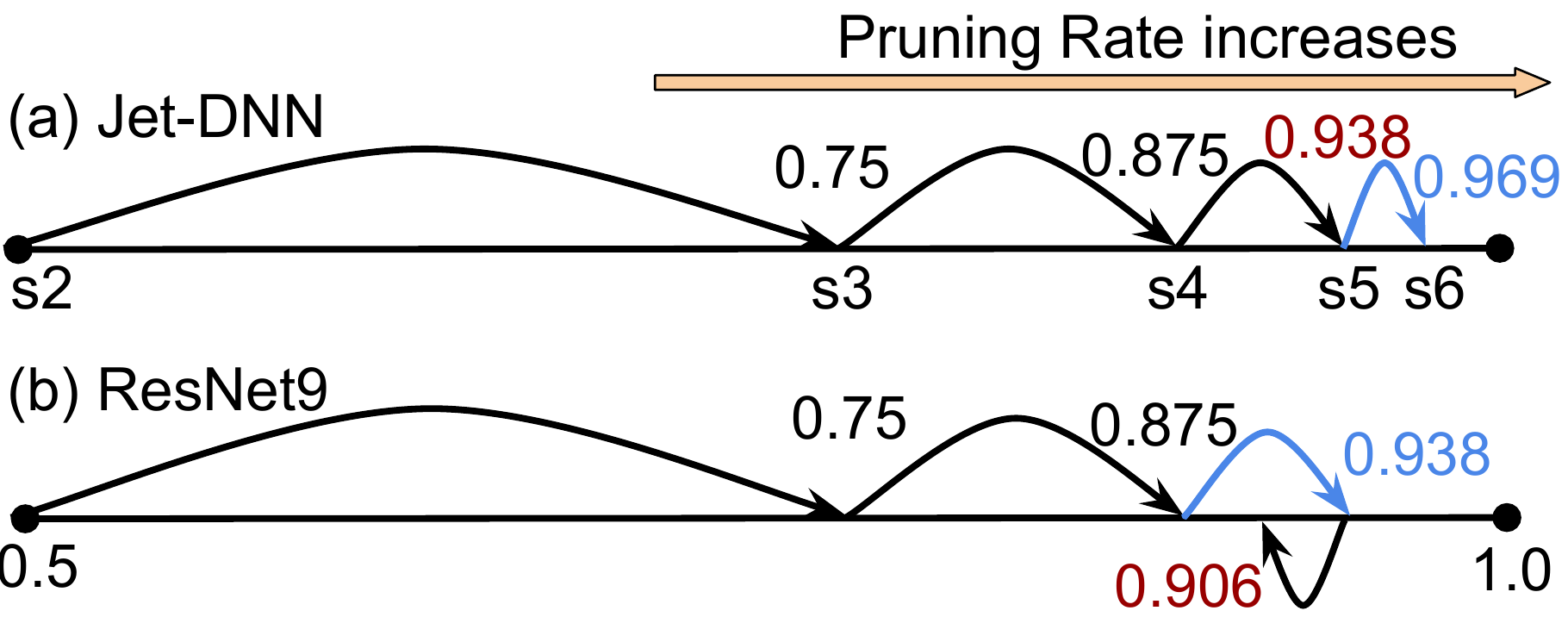}
\vspace{-0.3cm}
\end{center}
   \caption{The auto-pruning algorithm applied to models, (a) Jet-DNN and (b) ResNet9, with binary search direction shown. Omitting step s1 for visibility. The blue arrow indicates an accuracy loss $>$ user threshold; red denotes the optimal pruning rate. }
\label{fig:auto_pruning}

\end{figure}

Starting at 0\% pruning rate, the auto-pruning algorithm obtains initial accuracy
$Acc_{p0}$ at step~1 (s1).
It then uses a binary search approach, increasing or decreasing the pruning rate based on whether the accuracy loss is within a user-defined tolerance ($\leq \alpha_{p}$). The algorithm terminates when the rate difference is below a threshold ($\beta_{p}$). The number of steps is determined by $1+log_{2}(1/\beta_{p})$. Algorithm search steps and direction are shown in Fig.~\ref{fig:auto_pruning}.
Both the tolerance ($\leq \alpha_{p}$) and threshold ($\beta_{p}$) values are initially set to 2\% in this work.  

\begin{figure}
\begin{center}
\includegraphics[width=1.0\linewidth]{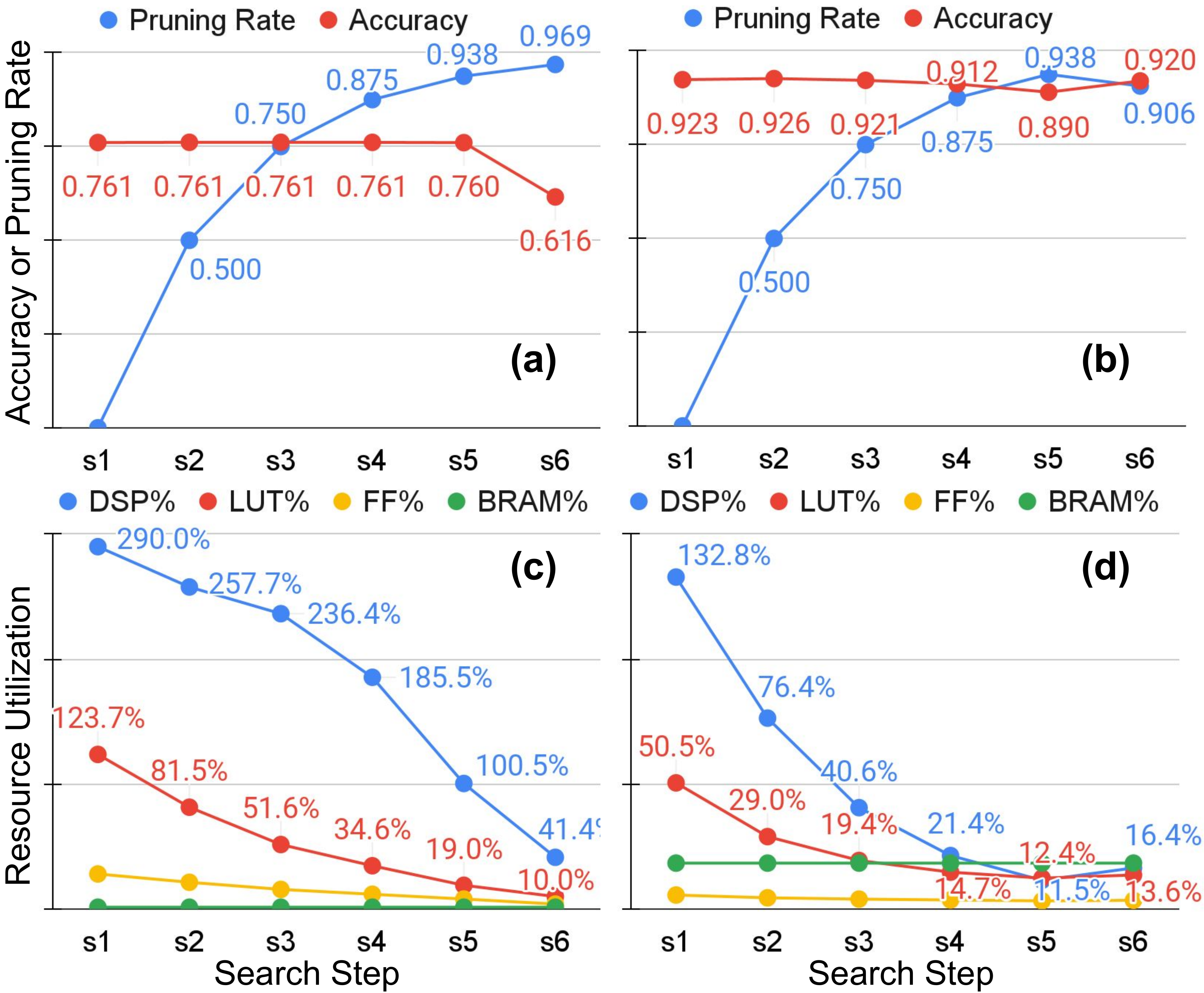}
\end{center}
\vspace{-0.1cm}
   \caption{(a) Pruning rates and accuracy of Jet-DNN.
  (b) Resource utilization of Jet-DNN design candidates with pruning on Xilinx Zynq 7020.
  (c) Pruning rates and accuracy of ResNet9.
  (d) Resource utilization of ResNet9 design candidates with pruning on Xilinx U250.}
\label{fig:strategy_results}
\vspace{0.1cm}
\end{figure}

The effectiveness of the auto-pruning algorithm is demonstrated in Fig.~\ref{fig:strategy_results}. Fig. 4(a) and (b) depict the pruning rate and accuracy for Jet-DNN and ResNet9 in each step, while Fig. 4(c) and (d) show the resources utilization. As the pruning rate increases, hardware resource requirements, particularly DSPs and LUTs, decrease, leading to improved FPGA performance. The design candidate with the highest pruning rate within the allowed tolerance is selected.

\textbf{Scaling strategy.} To accommodate a large DNN design on an FPGA, our framework supports the SCALING $O$-task which automatically reduces the layer size while tracking the accuracy loss $\alpha_s$. The search stops when the loss exceeds $\alpha_s$.  
This parameter can be adjusted to achieve further size reduction with minimal impact on accuracy. This work sets $\alpha_s$ to 0.05\%, which allows for model size reduction with negligible accuracy loss.

\textbf{Quantization strategy.} Our framework supports the QUANTIZATION $O$-task to automate mixed-precision quantization for networks. It operates at the HLS C++ level, providing more direct control over hardware optimizations and reducing unintended side effects when translating DNN models to HLS C++ using tools such as HLS4ML. 
The resulting precision configuration is directly instrumented into the C++ kernel, and a co-design simulation evaluates the accuracy of the quantized model. 
If the accuracy loss is within tolerance ($< \alpha_q$), this process is repeated. This work sets $\alpha_q$ to 1\%.

\begin{figure}
\begin{center}
\includegraphics[width=1.0\linewidth]{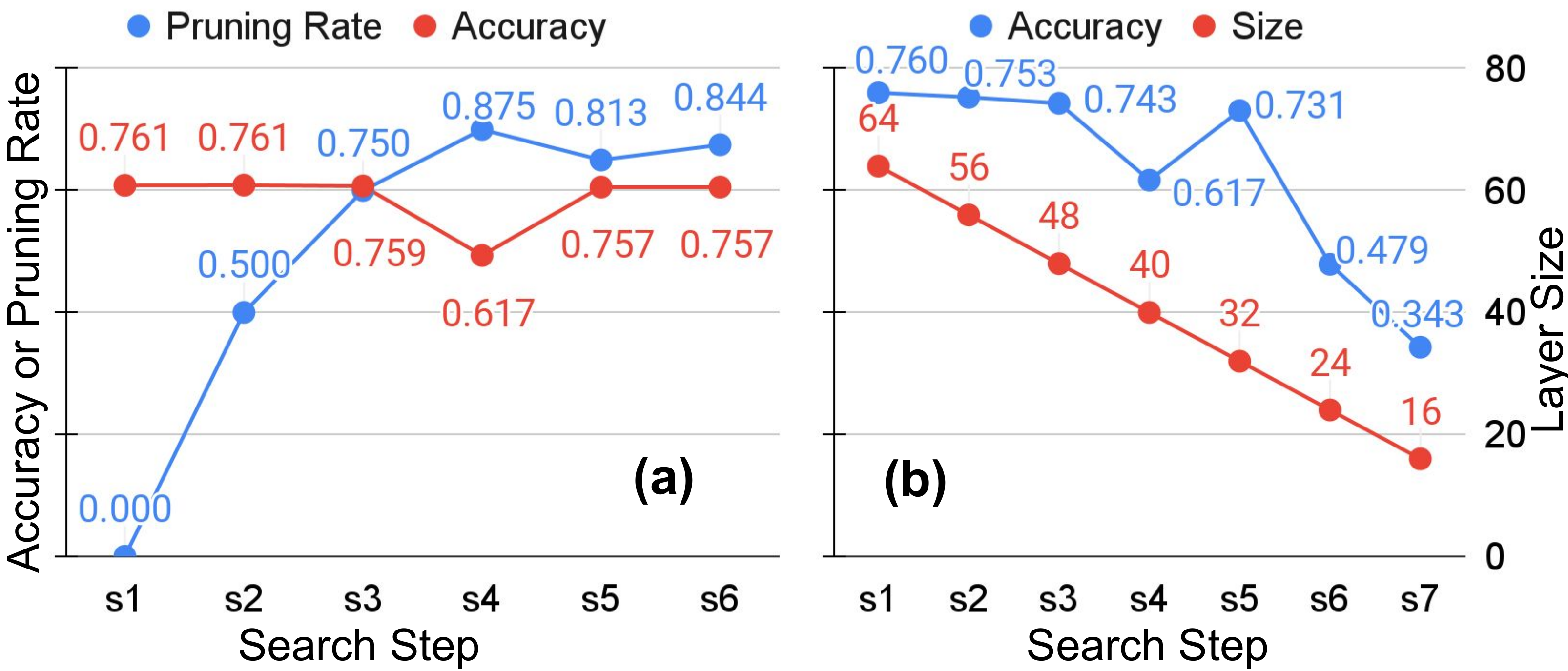}
\end{center}
\vspace{-0.1cm}
   \caption{
  (a) Jet-DNN accuracy and pruning rates with scaling then pruning.
  (b) Jet-DNN accuracy and layer size with pruning then scaling.}
\label{fig:combined_strategy_results}
\vspace{0.1cm}
\end{figure}

\textbf{Combining $O$-tasks.}
With our framework, new strategies can be derived by building and revising the design-flow architecture. For instance, by inserting the SCALING $O$-task before the PRUNING $O$-task in Fig.~\ref{fig:design_flow}(a), a custom combined strategy is created with results shown in Fig.~\ref{fig:combined_strategy_results}(a). The new optimal pruning rate is 84.4\%, lower than the previous 93.8\%, due to reduced redundancy from the preceding scaling task. By switching the order of the $O$-tasks, a different optimization is achieved, resulting in a 0.7\% accuracy drop after one scaling step, as seen in Fig.~\ref{fig:combined_strategy_results}(b). Moreover, the three optimization $O$-tasks, pruning, scaling, and quantization, can be integrated into a single automated cross-stage strategy to enhance both performance and hardware efficiency, as illustrated in Fig.~\ref{fig:design_flow}(b) and (c). 
\textbf{Discussion and Comparison.} 
Our evaluation results indicate that our combined $O$-task optimization strategy typically outperforms  single $O$-task techniques. Furthermore, the order in which these optimization techniques are applied plays a crucial role, as different orders produce varying final results. 


To highlight the advantages of our framework, we compared our design flow results to those of other studies targeting low-latency, low-resource, fully unfolded FPGA implementations, including original Jet-DNN~\cite{duarte2018fast} using HLS4ML, LogicNets~\cite{umuroglu2020logicnets}, QKeras-based Q6~\cite{coelho2021automatic}, and AutoQKeras-based QE and QB~\cite{coelho2021automatic}. All designs use the same architecture, except for JSC-L, which employs a larger architecture. Our "S+P+Q" design with $\alpha_q$ set to 0.01 achieves an accuracy of 74.1\% for JSC-S, outperforming JSC-M and JSC-L, which have accuracies of 70.6\% and 71.8\%, respectively. Compared to Q6, our design has 0.8\% higher accuracy, uses 2.5 times fewer DSP blocks, and 5.7 times fewer LUTs. Furthermore, our design outperforms both QE and QB, achieving over 3.3\% higher accuracy and lower resource usage. Our design also boasts lower latency than all Q6, QE, and QB designs, demonstrating the benefits of our framework. Inspired by the AutoQKeras design QB, which minimizes model bit consumption, we further optimize our design by tuning the parameters, such as $\alpha_q$. Increasing the quantization $O$-task's tolerant accuracy loss ($\alpha_q$) from 1\% to 4\% results in a smaller model with DSP usage 3 times lower than AutoQkeras' most efficient model, QE (see~\tabref{table:cmp_fpga}). 
Although the model accuracy decreases to 72.8\%, it remains higher than the optimized AutoQKeras designs QB and QE.

\input{tables/cmp_fpga03.tex}

%% file: tables/cmp_fpga03.tex
\begin{table}[t]
\centering

\caption{Performance comparison with the FPGA designs of Jet-DNN network using other approaches on Xilinx FPGAs
}
\label{table:cmp_fpga}
\scalebox{.80}{
\begin{threeparttable}
\centering
\begin{tabular}{c| c | c | c |c |c | c |c |c }
\toprule
 
Model 
& $\alpha_q$
& FPGA 
& \begin{tabular}[c]{@{}c@{}}Acc. \\(\%) \end{tabular}
& \begin{tabular}[c]{@{}c@{}}Lat. \\(ns) \end{tabular}
& \begin{tabular}[c]{@{}c@{}}Lat. \\(cycles) \end{tabular}
& \begin{tabular}[c]{@{}c@{}}DSP \\(\%) \end{tabular}
& \begin{tabular}[c]{@{}c@{}}LUT \\(\%) \end{tabular}
& \begin{tabular}[c]{@{}c@{}}Power \\(W) \end{tabular}
 \\

\midrule
\begin{tabular}[c]{@{}c@{}}HLS4ML \\Jet-DNN~\cite{duarte2018fast} \end{tabular}
& - & KU115 & 75 & 75 & 15 
& \begin{tabular}[c]{@{}c@{}}954 \\(17.3) \end{tabular}
& - 
& -   \\
\midrule

\begin{tabular}[c]{@{}c@{}}LogicNets \\JSC-M~\cite{umuroglu2020logicnets}   \end{tabular}
& - & VU9P & 70.6 & NA & NA & 0 (0) 
& \begin{tabular}[c]{@{}c@{}}14,428 \\(1.2) \end{tabular}
& - \\
\midrule
\begin{tabular}[c]{@{}c@{}}LogicNets \\JSC-L~\cite{umuroglu2020logicnets}   \end{tabular}
& - & VU9P & 71.8 & 13\tnote{a} & 5 & 0 (0) 
& \begin{tabular}[c]{@{}c@{}}37,931 \\(3.2) \end{tabular}
& - \\
\midrule

\begin{tabular}[c]{@{}c@{}}Qkeras \\Q6~\cite{coelho2021automatic} \end{tabular}
& - & VU9P & 74.8 & 55 & 11 
& \begin{tabular}[c]{@{}c@{}}124 \\(1.8) \end{tabular}
& \begin{tabular}[c]{@{}c@{}}39,782 \\(3.4) \end{tabular}
& - \\
\midrule

\begin{tabular}[c]{@{}c@{}}AutoQkeras \\QE~\cite{coelho2021automatic} \end{tabular}
& - & VU9P & 72.3 & 55 & 11 
& \begin{tabular}[c]{@{}c@{}}66 \\(1.0) \end{tabular}
& \begin{tabular}[c]{@{}c@{}}9,149 \\(0.8) \end{tabular}
& - \\
\midrule

\begin{tabular}[c]{@{}c@{}}AutoQkeras \\QB~\cite{coelho2021automatic} \end{tabular}
& - & VU9P & 71.9 & 70 & 14 
& \begin{tabular}[c]{@{}c@{}}69 \\(1.0) \end{tabular}
& \begin{tabular}[c]{@{}c@{}}11,193 \\(0.9) \end{tabular}
& - \\
\midrule
\midrule
\begin{tabular}[c]{@{}c@{}}This work \\(same to~\cite{duarte2018fast}) \end{tabular}
&  1\% & VU9P &  \textbf{76.1} & 70  & 14 
& \begin{tabular}[c]{@{}c@{}}638 \\(9.3) \end{tabular}
& \begin{tabular}[c]{@{}c@{}}69,751 \\(5.9) \end{tabular}
& 2.51~\tnote{b}  \\
\midrule
\begin{tabular}[c]{@{}c@{}}This work \\S$\rightarrow$P$\rightarrow$Q \end{tabular}
&  1\% & VU9P &  \textbf{75.6} &45 & 9 
& \begin{tabular}[c]{@{}c@{}}50 \\(0.7) \end{tabular}
& \begin{tabular}[c]{@{}c@{}}6,698 \\(0.6) \end{tabular}

& 0.199~\tnote{b}  \\
\midrule
\begin{tabular}[c]{@{}c@{}}This work \\S$\rightarrow$P$\rightarrow$Q \end{tabular} 
& 4\% & VU9P & \textbf{72.8} & 40 & 8 
& \begin{tabular}[c]{@{}c@{}}23 \\(0.2) \end{tabular}
& \begin{tabular}[c]{@{}c@{}}7,224 \\(0.6) \end{tabular}

& 0.166~\tnote{b}  \\

\bottomrule
\end{tabular}

   \begin{tablenotes}
    \footnotesize
     \item[a] A clock frequency of 384 MHz is used and the final softmax layer is removed. 
     \item[b] Dynamic power reported by Vivado. The static power is about 2.5W for all these designs.  
    \normalsize
   \end{tablenotes}
\end{threeparttable}}
\end{table}

%% file: sections/conclusion.tex
\section{Conclusion}
\label{sec:conclusion}

This paper presents a novel co-optimization framework for FPGA-based DNN accelerators, enabling efficient development of customized cross-stage design flows. 
Our approach reduces DSP resource usage by up to 92\% and LUT usage by up to 89\% while maintaining accuracy. Compared to existing research, our approach achieves higher model accuracy and uses 3 times fewer DSP resources, highlighting the benefits of our framework. 
Future work includes exploring more efficient search techniques with larger numbers of $O$-tasks, introducing control tasks to cover bottom-up and parallel flows, supporting more types of neural networks such as graph~\cite{que2022ll} and transformer~\cite{wojcicki2022accelerating} networks, 
and utilizing new FPGA resources such as AI Engines~\cite{xilinx_white} and AI Tensor Blocks~\cite{langhammer2021stratix}.

\section*{Acknowledgement}
The support of the United Kingdom EPSRC (grant numbers EP/V028251/1, EP/L016796/1, EP/N031768/1, EP/P010040/1, and EP/S030069/1), CERN, AMD and SRC is gratefully acknowledged.